# TEXT-TO-SPEECH CONVERSION WITH NEURAL NETWORKS: A RECURRENT TDNN APPROACH

*O. Karaali, G. Corrigan, I. Gerson, and N. Massey*
Speech Processing Laboratory
Motorola, Inc.
1301 E. Algonquin Rd., Schaumburg, IL 60196, U.S.A.

## ABSTRACT

This paper describes the design of a neural network that performs the phonetic-to-acoustic mapping in a speech synthesis system. The use of a time-domain neural network architecture limits discontinuities that occur at phone boundaries. Recurrent data input also helps smooth the output parameter tracks. Independent testing has demonstrated that the voice quality produced by this system compares favorably with speech from existing commercial text-to-speech systems.

## 1. INTRODUCTION

The notion of using a neural network, or other machine learning system, to implement components in a text-to-speech system is an attractive one. A system trained on actual speech may learn subtler nuances of variation in speech than can presently be incorporated into rule-based or concatenation text-to-speech systems. The data storage requirements are also an order of magnitude smaller for a well-designed neural network than for a concatenation system. It should also be easier to train a neural network on a new language than to determine a rule set for that language. Training the network might even be easier than identifying and extracting the concatenation units necessary for a new language.

Several attempts have been made to implement various components of a text-to-speech system with neural networks, including several that implemented the phonetic component [1], [2], [3], [4], [5]. This is the component that converts a phonetic description of an utterance, including segment durations for each phone, into a series of acoustic descriptions of frames of speech. Most of these prior attempts to use neural networks for phonetic components described the phonetic context of each speech frame using input sets that represented the current phoneme, and one or more preceding or following phonemes, and extra inputs indicating the position of the current acoustic frame in the current phoneme. When two adjacent acoustic frames are in different phonetic segments, all of the phoneme representations change between the two frames. These input discontinuities are reflected with large discontinuities in the output data, which are heard as warbling in the generated speech.

This paper describes the phonetic component of a text-to-speech system which uses a recurrent time-delay neural network (TDNN) approach to generate high-quality speech.

## 2. SYSTEM OVERVIEW

The complete system is shown in Figure 1. The text-to-speech system includes a text-to-linguistic description subsystem, a neural network used to assign a duration to each phonetic segment, a neural network used to convert the linguistic description into a series of coder parameter vectors, and the synthesis section of a speech coder.

The text-to-linguistic description subsystem produces a description of the speech to be generated that includes a sequence of phones along with prosodic and syntactic annotations. The details of this subsystem will not be described here, except to note that it generates the same marks that are used to label the database, described below, and that the timing of the marked events is established relative to the timing of the phonetic segments, so that determining the segment durations determines the timing of the other marked events.

The segment durations are also computed using a neural network. This is described in [6]. The speech coder parameter set is described in the discussion of training data below.

## 3. TRAINING DATA

In order to train a neural network to perform the phonetic-to-acoustic mapping, it was necessary to prepare an appropriate database. This database, consisting of a set of recordings of speech from a single speaker, was then labeled phonetically, syntactically, and prosodically. The recordings were processed by the analysis portion of a parametric vocoder, to produce a series of coder parameter vectors describing the acoustic characteristics of 5 ms. frames of speech. The speech labels were also processed to generate neural network input vectors describing the phonetic and prosodic context of the 5ms. speech frames. The neural network was trained to generate an appropriate coder parameter vector in response to each neural network input vector.

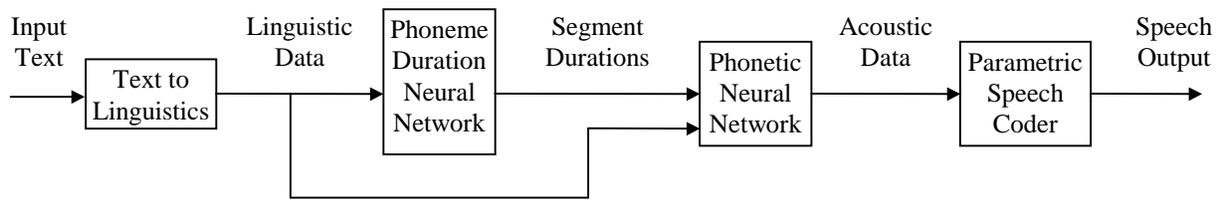

**Figure 1:** Text-to-speech system

The steps in generating these training vectors is described in more detail below.

### 3.1. Speech Recordings

In order to make it feasible for the network to learn the phonetic-to acoustic mapping, a single speaker was used for all of the speech recordings. This speaker is a male speaker from the Chicago area. The principle portion of the database, a collection of 480 sentences from the Harvard sentence lists, was recorded when the speaker was 36 years old. Additional recordings were made two years later in order to increase the prosodic variety of the recorded speech. These recordings included questions, isolated words, paragraph-length materials, and selections from dramatic works. The recordings were made in a soundproof room with a close-talking microphone.

### 3.2. Speech Labeling

The speech was phonetically labeled in the same manner as the TIMIT database. In order to allow the neural network to learn the phonetic-to-acoustic mapping, additional information was provided by marking syllable, word, phrase, and clause boundaries, tagging each word as a content or function word, and marking syllables with primary or secondary stress.

### 3.3. Voice Coder

Much of the current research in speech coding uses a source-filter model, and models the source using a codebook of excitation vectors. These codebook approaches are inappropriate for use in neural network speech synthesis. The codebooks are typically quite large, and would require individual outputs to select each codebook entry. This would make the neural network unwieldy. Binary target values of any kind may also lead to problems when mixed with continuous targets. A coder that uses continuous parameter vectors is, therefore, desirable.

The coder design used for this system uses a source-filter model. The filter is an autoregressive filter, using line spectral frequencies to describe the filter. A mixed-source excitation model [7] was used. In this model, excitation consists of a low-frequency band of periodic excitation and a high-frequency band of aperiodic excitation. The parameters used to describe the source were the energy of the speech signal, the pitch of the periodic excitation, and the boundary frequency between the bands.

### 3.4. Input Processing

The purpose of the input processing is to provide the information contained in the speech labels to the neural network in an appropriate format. Previous attempts to use neural networks in speech synthesis have represented the phonetic context for each frame by having input representing the phonetic segment containing the frame, the surrounding segments, and the position of the frame within the segment. The problem with this representation is that all of the input changes at one time at each segment boundary. This can produce significant discontinuities in the neural network output at these points. These discontinuities result in audible artifacts in the generated speech.

A Time-Delay Neural Network (TDNN) does not have this problem. Figure 2 illustrates the TDNN input structure. For each 5 ms. frame of speech, there is an input identifying the phone to be produced during the phonetic segment containing the frame. The input to the neural network includes inputs describing the phonetic segment associated with a number of surrounding frames. Only a few of these inputs should change between any two frames, so the size of discontinuities in the output is reduced.

The TDNN window introduces some new problems. The amount of context used in computing the coder parameters for a frame is determined by the width of the TDNN window. Increasing number of frames sampled in the TDNN window, however, increases the network size; a wide enough TDNN window may make the network unwieldy. This problem can be alleviated by non-uniform sampling of the TDNN window. Near the center of the window, every frame is sampled to provide an input to the neural network. Near the edges, the phonetic data is sampled less often.

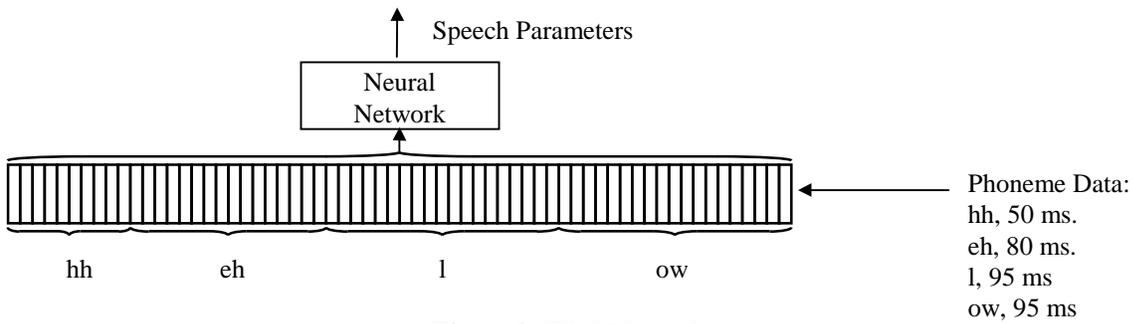

**Figure 2:** TDNN Input Structure

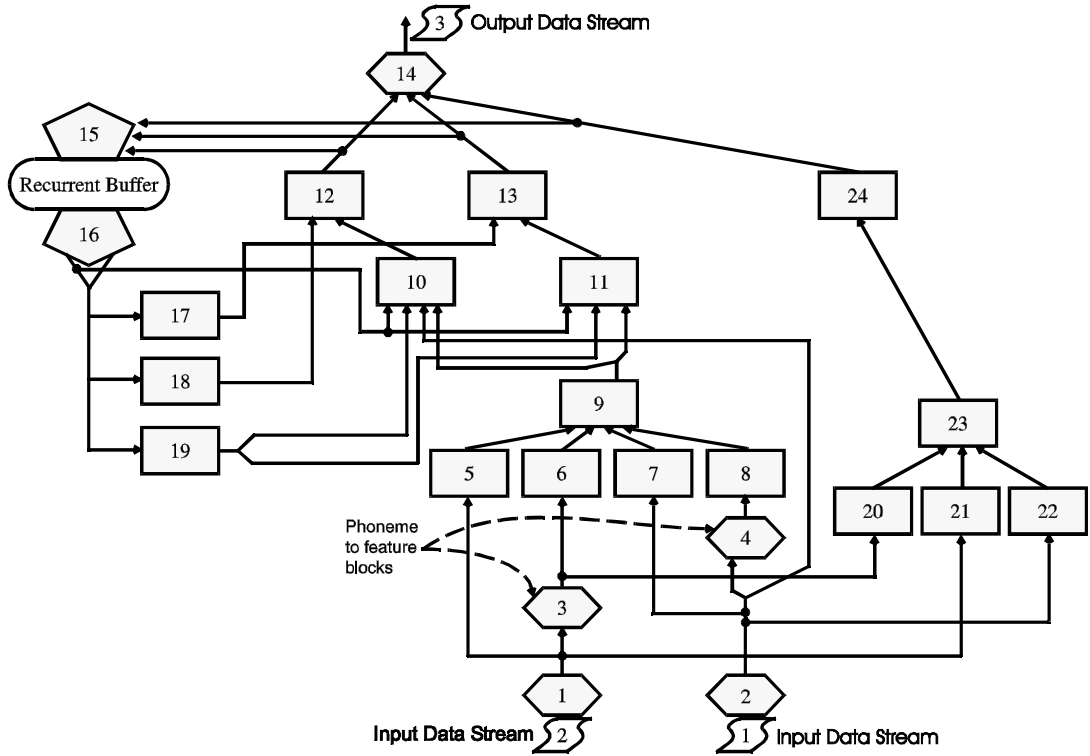

**Figure 3: Neural Network Architecture**

In addition to the 5 ms. TDNN coding of the phoneme labels and phoneme features, the duration and distance from the current frame of the current phonetic segment, the four preceding segments, and the four following segments are coded into the network. This coding provides more context information and becomes especially useful when a string of long phonemes occupies the TDNN window. The TDNN window is 300 ms. wide, and if there are a few very long phonemes, the TDNN window will not be able to cover much context outside of these long phonemes. The duration and distance coding of nine phonemes insures that a context of this size is always available to the network.

In order to provide some domain-specific knowledge to the neural network, the phonemes were encoded not only as a one-of-n binary vector, but as a vector of articulatory features. This redundant information enhances the network's ability to learn similarities between phonemes.

In order to generate intonation, the labeled syntax and stress information was also provided to the neural network. Syllable and word characteristics were encoded using a TDNN representation, while duration and distance coding was used to mark the boundaries of syntactic elements such as phrases and clauses.

## 4. NETWORK ARCHITECTURE

The phonetic neural network system is composed of multiple neural network subsystems as shown in Figure 3, where all the rectangles are neural network blocks. The blocks 1, 2, and 14 are the I/O blocks and provide an interface to data streams 2, 1, and 3. Blocks 3 and 4 convert phoneme labels to phoneme features. Blocks 15 and 16 provide the recurrent buffers for feedback paths. Blocks 17, 18, and 19 transform the feedback information to a form more acceptable to their target blocks. Block 5 works on TDNN phoneme labels. Block 6 works on TDNN phoneme features. Block 7 works on duration and distance of phoneme labels. Block 8 works on duration and distance of phoneme features. Blocks 9, 10, 11, 12, and 13 provide higher level neural network processing. Blocks 20, 21, 22, 23, and 24 generate 10 bands of power spectrum where the band boundary frequencies are chosen according to the formant boundaries of the speaker.

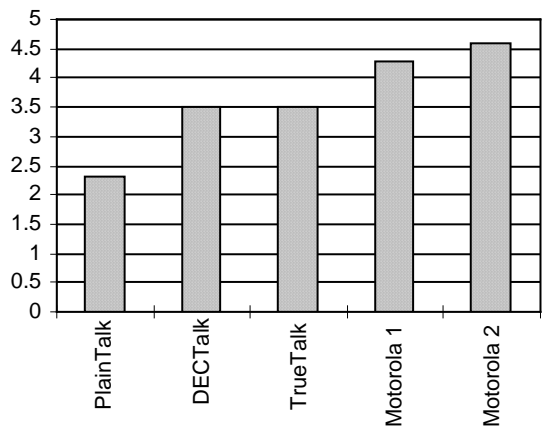

**Figure 4:** Mean Opinion Scores

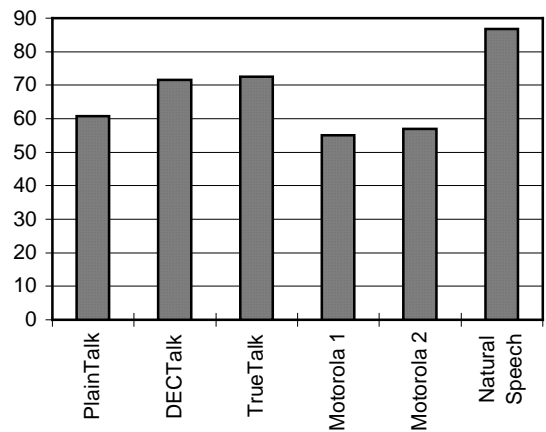

**Figure 5:** Percent of Words Recognized

## 5. EXPERIMENTAL RESULTS

The performance of this neural network based system was compared to existing systems by an independent tester [8]. Two evaluations were performed. The first provided sentence-length materials to listeners, who judged the acceptability of the speech on a scale of one to seven, with seven being most acceptable and one least acceptable. The results of this evaluation are shown in Figure 4. Motorola 1 used phone durations generated by a neural network, as described in [6], while Motorola 2 used durations from natural speech. Both Motorola systems performed significantly better than the other systems.

Figure 5 shows the results of a segmental intelligibility experiment. In this experiment, subjects were asked to identify isolated monosyllabic words from the tested systems. Figure 4 illustrates the percent of the words the subjects identified correctly. In this experiment, the Motorola systems, which had not yet been trained on isolated words, did not perform as well as some other systems.

## 6. CONCLUSION

This paper presents the design of a recurrent TDNN which performs the phonetic-to-acoustic mapping in a text-to-speech system. A complete system using this neural network has been implemented on a personal computer and runs in real time. The system performs competitively with existing commercial text-to-speech systems, as demonstrated in independent tests.